 \documentclass[pmlr,twocolumn,10pt]{jmlr} 

\usepackage{booktabs}
\usepackage{siunitx}

\usepackage[switch]{lineno}

\usepackage{subfiles}

\theorembodyfont{\upshape}
\theoremheaderfont{\scshape}
\theorempostheader{:}
\theoremsep{\newline}

\jmlrvolume{LEAVE UNSET}
\jmlryear{2024}
\jmlrsubmitted{LEAVE UNSET}
\jmlrpublished{LEAVE UNSET}
\jmlrworkshop{Machine Learning for Health (ML4H) 2024} 

\title[In-Context Learning for Preserving Patient Privacy]{In-Context Learning for Preserving Patient Privacy: \\ A Framework for Synthesizing Realistic Patient Portal Messages}

\author{%
\Name{Joseph Gatto} \Email{joseph.m.gatto.gr@dartmouth.edu}
\AND
\Name{Parker Seegmiller} \Email{matthew.p.seegmiller.gr@dartmouth.edu}
\AND
\Name{Timothy E. Burdick} \Email{timothy.e.burdick@hitchcock.org }
\AND
\Name{Sarah M. Preum} \Email{sarah.m.preum@dartmouth.edu}
}
\begin{document}

\maketitle

\begin{abstract}
Since the COVID-19 pandemic, clinicians have seen a large and sustained influx in patient portal messages, significantly contributing to clinician burnout. To the best of our knowledge, there are no large-scale public patient portal messages corpora researchers can use to build tools to optimize clinician portal workflows. Informed by our ongoing work with a regional hospital, this study introduces an LLM-powered framework for configurable and realistic patient portal message generation. Our approach leverages few-shot grounded text generation, requiring only a small number of de-identified patient portal messages to help LLMs better match the true style and tone of real data. Clinical experts in our team deem this framework as HIPAA-friendly, unlike existing privacy-preserving approaches to synthetic text generation which cannot \textit{guarantee} all sensitive attributes will be protected. Through extensive quantitative and human evaluation, we show that our framework produces data of higher quality than comparable generation methods as well as all related datasets. We believe this work provides a path forward for (i) the release of large-scale synthetic patient message datasets that are stylistically similar to ground-truth samples and (ii) HIPAA-friendly data generation which requires minimal human de-identification efforts. 

\end{abstract}


\paragraph*{Data and Code Availability} All code and data from this study which does not include sensitive patient information can be found at \url{https://github.com/Persist-Lab/SyntheticPortalGen}.

\paragraph*{Institutional Review Board (IRB)}
This study was approved by the Dartmouth College Institutional Review Board (IRB Number MOD00018812)

\section{Introduction}
\label{sec:intro}

\textit{Have you sent a message to your doctor recently?} Because in the past few years, there has been a significant increase in use of patient-facing healthcare applications which allow patients to send textual messages to their provider \cite{doi:10.1177/1357633X221146810}. Electronic health record (EHR) applications such as Epic's MyChart, for example, reportedly had 200 million active users as of early 2021, with 80 million joining within the previous 12 months \cite{epic2024}. Unfortunately, this shift in doctor-patient communication paradigms has contributed significantly to doctor burnout as clinicians have not been provided with additional resources to handle the portal message surge \cite{death_by_portal}. To help reduce clinician workload, recent studies have explored the use of AI tools to optimize cognitively demanding tasks such as portal message triage \cite{pmlr-v126-si20a, gatto2022identify, 10.1001/jamanetworkopen.2023.22299}, routing, \cite{Harzand2023.11.27.23298910}, and response writing \cite{info:doi/10.2196/46939, Kozaily2023.09.12.23295452, ATHAVALE2023100019}. However, most of these studies use sensitive data that cannot be released to the community. Some prior work have explored patient messaging through the lens of data sourced from public medical Q\&A forums \cite{gatto2022identify}. However, such platforms operate assuming that the physician who will respond is unfamiliar with the patient. This produces patient messages that are highly dissimilar to actual portal message data, where personal relationships with clinicians and historical EHR data lead to patient messages that are stylistically different, often containing implicit references to EHR data and prior encounters with providers. See Appendix \ref{apd:A} for an illustrating example of this issue.

Given the sensitive nature of patient message data, recent advancements in language modeling may be useful in generating realistic synthetic datasets which can be made public. For example, one could fine-tune a language model on data from the true distribution and then generate realistic synthetic data. Unfortunately, this paradigm may generate samples that leak sensitive patient information learned during training. Even privacy-preserving mechanisms such as Differential Privacy (DP) \cite{dp-transformers} cannot provide a risk-free guarantee that no sensitive patient attributes would be released during generation. This is because DP language models are trained not to leak full training data instances, but there remains a risk of generating a sensitive token (e.g., a patient's last name) violating the patient's privacy. Thus, a solution to this problem must generate data that reflects true patient portal conversational style, semantics, and structure, all while \textit{ensuring synthetic samples pose zero risk of protected health information (PHI) leakage}. 

This issue can perhaps be solved via LLM prompting. LLMs have tremendous capacity to generate texts while following a set of instructions \cite{10.1162/coli_a_00523}, potentially removing the need for sensitive training data. One could thus prompt an LLM to generate a patient portal message containing a pre-defined set of details. Unfortunately, as discussed in prior work \cite{wang2024largelanguagemodelsreplace, gupta2024biasrunsdeepimplicit, liu2024evaluatinglargelanguagemodel}, LLMs are biased, highly-formal, and struggle to capture the natural voice of people from different identity groups. Thus, off-the-shelf applications of LLMs to this task are ill-suited to generate data that matches real patient message style. 

To address these challenges, we propose PortalGen, a two-stage, HIPAA-friendly, LLM-powered framework for the configurable generation of realistic patient message data. In stage 1, PortalGen uses few-shot prompting of LLMs to transform codes from healthcare databases into portal message prompts. This provides a means of generating diverse large-scale message corpora covering a wide variety of health situations. In stage 2, we use grounded generation with a small number of de-identified patient messages to convert prompts from stage-1 into patient messages. Grounded generation \cite{veselovsky2023generating} is a technique that includes samples from the target distribution in the prompt, encouraging LLM outputs to be more stylistically and semantically faithful to the nuances of real training samples. PortalGen performs grounding with just 10 de-identified patient messages, providing a framework for researchers and institutions to release realistic synthetic patient portal message data without requiring large-scale de-identification efforts. 
Our results demonstrate that PortalGen produces data that is highly similar to real data, outperforming baseline data synthesis techniques and showing strong contrast with related public medical Q\&A datasets.
\section{Dataset Overview}
\label{sec:data}

Our patient portal messages are sourced from a large-scale dataset of 610k patient messages collected from a large academic medical system in the United States. All messages were sent between 1/2020 - 9/2024. The dataset contains 10,526 unique patients, where the population is 29\% male, 51\% female, and 20\% un-known gender, with ages between 17 - 81.

The PortalGen framework leverages only 10 real patient message examples. We choose n=10 as it provides sufficient context to the LLM while minimizing the de-identification efforts required for community data release. The authors manually selected 10 representative messages with varying lengths and health scenarios. Each message was manually de-identified by a human annotator who has been trained on how to handle sensitive patient data and how to de-identify portal messages. All removed PHI elements were randomly replaced using viable substitutes for each PHI category (e.g. ``My name is Jane" $\rightarrow$ ``My name is Victoria"). All other data used for training and evaluation of PortalGen is sampled from the remaining population of patient messages.

\section{Methods}
\label{sec:methods}

In this section, we describe our proposed two-stage framework, PortalGen. 

\noindent \textbf{Stage \# 1: } In effort to generate a large number of patient portal messages covering a diverse range of health conditions, we develop an LLM-driven framework for converting ICD-9 codes into message prompts. ICD-9 codes are a widely used standardized mapping of numeric codes to health conditions. Our framework allows us to use public ICD-9 databases \footnote{https://github.com/kshedden/icd9} to source various health conditions and convert them into patient portal message prompts. Each ICD-9 code in the database comes with a brief description, which we use as the input in our framework. 

We map ICD-9 descriptions to portal message prompts using GPT-3.5 \footnote{gpt-3.5-turbo-instruct} with few-shot learning (k=4). To illustrate, consider an example we provide in context. (\textbf{ICD-9 Code Description:} Shoulder joint replacement $\rightarrow$ \textbf{Message Prompt:} Patient heard a snap while trying to lift heavy boxes after shoulder surgery, and is experiencing pain.) As you can see, we instruct the LLM to extrapolate the ICD-9 description into a realistic situation that could be the context of a patient message for someone with this ICD-9 code in their medical chart. We use GPT-3.5 due to it's popularity, low cost, instruction following ability, and fast inference speeds. The full prompt used in Stage \#1 as well as additional details can be found in Appendix \ref{apn:c}. 

We use the described framework to generate 1000 message prompts using descriptions from 1000 randomly sampled ICD-9 codes, which are used throughout the remainder of this study. 

\noindent \textbf{Stage \# 2: } To transform patient message prompts into synthetic patient message data, we employ grounded generation \cite{veselovsky2023generating}. Specifically, we include 10 real de-identified patient portal messages in the prompt when asking the model to synthesize a new message. This guides the LLM towards matching the style and prose of samples from the target distribution. For each real sample, we create a ground truth prompt that the LLM can use to learn how to map prompts to messages. This allows us to formulate our synthetic data generation as a 10-shot prompt, using 10 (prompt, message) pairs as in-context examples. The prompt used for Stage \# 2 can be found in Appendix \ref{apn:B}. 

\begin{figure}[!t]
    \centering
    \includegraphics[width=\columnwidth]{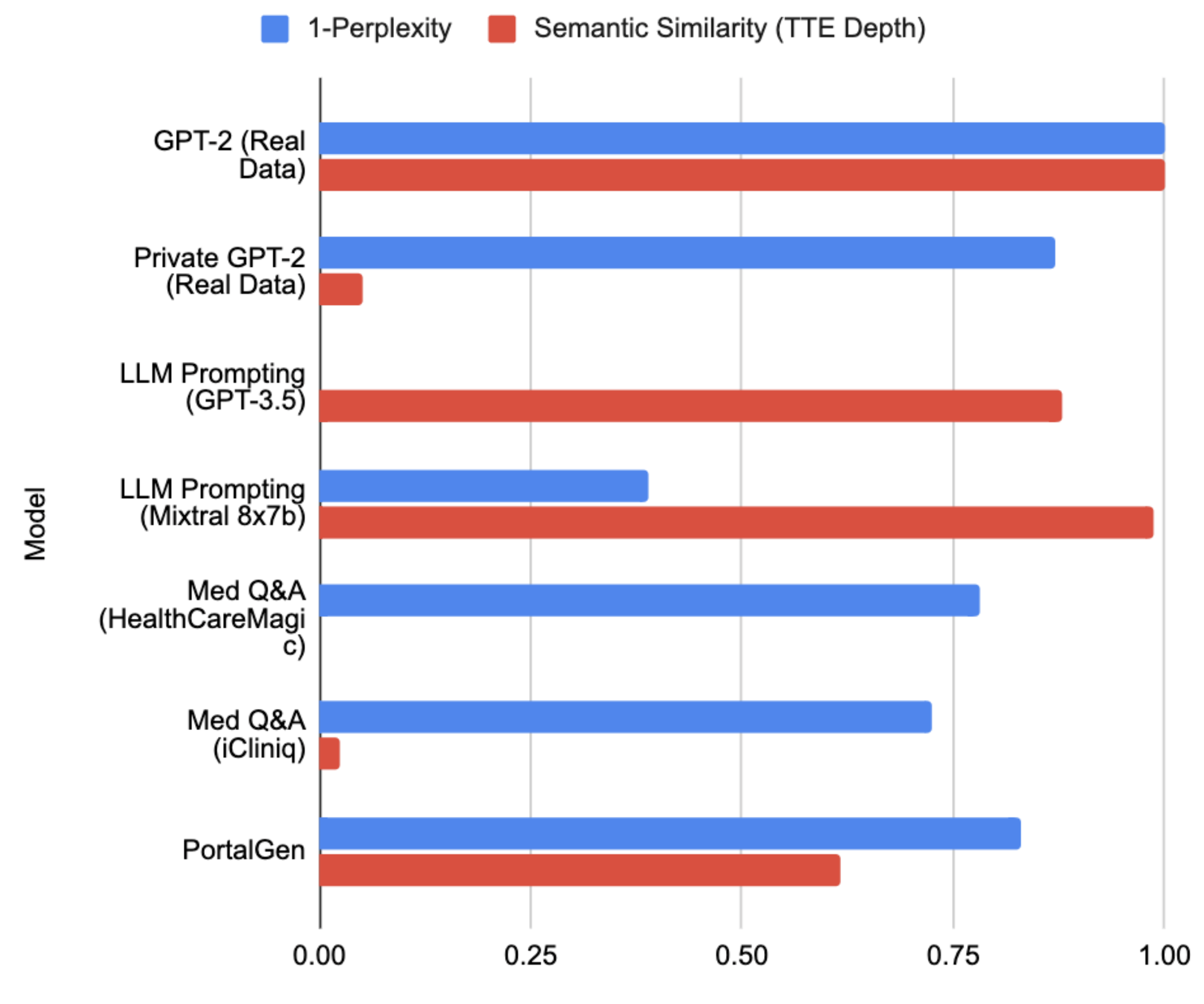}
    \caption{Our experiments show that PortalGen produces data which achieves the best balance of perplexity and semantic similarity amongst HIPAA-friendly generation methods. Note that all metrics are scaled between 0-1 (higher=better) for ease of comparison. }
    \label{fig:quant_results}
\end{figure}\vspace{-10pt}

\section{Experimental Setup}
We compare PortalGen to a variety of relevant baselines. 

\noindent \textbf{GPT-2 Fine-Tuned on Real Patient Data:} We fine-tune GPT-2 on 1,000 real patient portal messages as a ground truth baseline. We then generate 1,000 messages using this model. This experiment showcases performance when privacy preservation is not a concern. 

\noindent \textbf{GPT-2 Fine-Tuned with Differential Privacy:} We fine-tune GPT-2 with differential privacy \cite{dp-transformers} on 1,000 real patient portal messages. Due to the sensitive nature of this data type, we use a privacy budget of $\epsilon = 1$. We then generate 1,000 messages using this model. This experiment compares our approach to an industry-standard privacy-preserving generation technique. 

\noindent  \textbf{Zero-Shot LLM Prompting: } We generate 1,000 synthetic messages using our ICD-9-based prompts with two different LLMs, GPT-3.5-Turbo and Mixtral 8x7b \cite{jiang2024mixtralexperts}. This experiment showcases data quality without grounding. \\
\noindent  \textbf{PortalGen: } We evaluate our framework, PortalGen, on Mixtral 8x7b. Note that this study was performed on a secure computing platform with limited GPU resources and no access to the internet, limiting our ability to explore many larger API-based LLMs. Mixtral 8x7b is the largest LLM we are able to explore in this study. Additional relevant configurations and parameters for all LLM-based experiments can be found in Appendix \ref{apn:B}. \\
\noindent  \textbf{Baseline Datasets: } We additionally compare our synthetic data to two open-source corpora sourced from online medical Q\&A forums, ICliniq.com \cite{huggingface_icliniq}  and HealthCareMagic.com \cite{huggingface_hcm}. We explore the semantic and stylistic differences between our synthetic corpora and 1,000 randomly sampled messages from each of these two datasets.

\subsection{Evaluation}
We explore three different evaluation metrics in this study. \textbf{(i) Perplexity Analysis: } We train a GPT-2 model from scratch using a given synthetic dataset. We then compute the mean perplexity of this model on a hold-out test set of 5,000 real patient messages. This experiment tells us whether an LLM trained with synthetic data is confused or surprised by real patient messages. We report 1 - the normalized perplexity in Figure \ref{fig:quant_results} for ease of comparison across metrics. \textbf{(ii) Semantic Similarity: }
We use statistical depth \cite{seegmiller-2023-tte-depth}, a measure of how semantically similar two corpora are, to determine if the synthetic dataset is semantically similar to our set of 5,000 test messages. This experiment tells us if the semantic structure of our synthetic data is similar to real messages. We report the normalized statistical depth Q-value (higher=more similar) in Figure \ref{fig:quant_results}. \textbf{(iii) Human evaluation: } We additionally have two human annotators with significant experience reviewing patient portal messages perform qualitative evaluation. Specifically, we randomly select 10 prompts used in the LLM-based experiments and have each annotator blindly rank the outputs from GPT-3.5, Mixtral 8x7b, and PortalGen from best to worst based on their interpretation of how the messages read in terms of stylistic and semantic similarity to real patient messages. We report the mean ranking of each model after this analysis.

\section{Results}

\textbf{Quantitative Evaluation: }In Figure \ref{fig:quant_results} we show the results of our quantitative experiments. On perplexity, we find that as expected, the two models trained on real patient message data achieve the best performance. When we compare PortalGen to LLM Prompting, we find a significant difference in perplexity, with LLM-prompting achieving the worst score across all experiments. This result provides evidence for our claim that LLMs without grounding struggle to generate data which stylistically and/or structurally matches real patient data. Additionally, we find that PortalGen data shows better perplexity than both Medical Q\&A datasets, highlighting the difference between the two data sources. 

When comparing each corpus using statistical depth, we again find that the GPT-2 model trained on real data achieves the highest score as expected. However, on this metric the privacy preserving generation begins to fail, as there is a known trade-off in generated content quality when training with privacy-preserving algorithms \cite{dp-transformers}. We find that LLM-prompting scores second highest on statistical depth, with PortalGen achieving the third highest score. This follows our intuition that LLMs mostly struggle to mimic style and prose, not following instructions to include certain pre-defined semantic details. Interestingly, both Med Q\&A datasets have extremely low semantic similarity compared to the ground truth data. This again provides quantitative evidence for the stark difference in medical Q\&A forum data and real patient messages. In summary, we find that PortalGen achieves the best balance of depth and perplexity across all generation strategies. \\
\textbf{Qualitative Evaluation:} Our human evaluation tasked annotators with blindly ranking the outputs (1=best, 3=worst) from three different generation strategies. We find that PortalGen outputs had an average rank of 1.55, GPT-3.5 had an average rank of 2.2, and Mixtral-8x7b had an average rank of 2.25. This indicates that \textbf{human annotators most often find PortalGen to generate the most realistic patient portal messages}. Qualitatively, we note that the gap between PortalGen performance and baseline LLMs is smaller when the prompt is more vague. E.g. a lower ranking PortalGen output has a prompt containing ``Patient is seeking advice on managing symptoms and coping mechanisms...". Conversely, one of the unanimous high ranking PortalGen outputs contains a more detailed prompt ``Patient is experiencing swelling and stiffness...". We share an example model output in Appendix \ref{apn:A_2}. 

\section{Discussion, Limitations, and Future Work}
\label{sec:discussion}

In this paper, we propose the use of LLM-grounded generation for the synthesis of patient portal messages. Our framework provides researchers with access to portal messages a realistic path towards large-scale data release. Our framework promotes the generation of diverse and realistic messages that may be used in the development of patient portal tools which optimize clinical workflows and reduce clinician burnout. A limitation of this study is that we cannot evaluate our framework on LLMs larger than 50 billion parameters due to the computational restrictions of our HIPAA-compliant workstation. Additionally, our work is limited in that we only explore portal messages from one healthcare system. Future works in this space may explore varying the number of samples used for grounding, as well as how to generate realistic message data without reliance on in-domain samples.

\bibliography{main}

\appendix

\section{Example Data}\label{apd:A}

\subsection{Comparing Patient Portal Messages to Online Medical Q\&A Data }
In this section, we provide an illustrating example of how real patient portal message data differs from related public datasets. First, consider the following example from the HealthCareMagic \cite{huggingface_hcm} dataset: 

\begin{quote}
    I am a 35 year old male and own a landscape business for 2 years, for the last 6 weeks It has become difficult for me to get through half the day without becoming exhausted (low energy that comes on fairly sudden, slightly shakey, foggy mind kinda like im in a dream) I have cut my hours back and have had employees make up the difference, thinking maby I am working to hard, energy still cuts out half way thru the day. I have had some life changes (seperation, moving to new house, extra debt to support expanding business) but nothing I would consider unmanageable. Any advice? I have setup a health exam with my doctor in September? Could this be physical or phsychological? Any advice from past experiencees? Thanks!
\end{quote}

We notice how the patient provides demographic information (i.e. age) in the body of the message, which does not typically occur in real data as there is an implicit understanding that the clinician has the patients medical chart available when reading their message. One can imagine how using NLP to classify how urgent or worrisome this patient's symptoms are, for example, may heavily depend on demographics, as fatigue may be considered more or less expected depending on the patients age. Developing a synthetic dataset for use with real messages must reflect the fact these portal message classification tasks are often multi-modal with many pieces of information needing to be pulled from the EHR. Additionally, the style and tone of this message reflect the fact that there is no relationship between the patient  and provider, as the message contains significant details about the patient's personal and professional life. Conversely, real patient message data often contains personal anecdotes, references to past conversations, and friendly language given the personal relationship the patient has with their provider. Such style and prose should be reflected in synthetic datasets. 

Now consider a synthetic patient portal message: 

\begin{quote}
    Hey Dr. James, 

    The pain is back again. I don't know what the next steps should be but the pain medication you gave me is NOT working. Is there something else I can try that you can call in or should I just schedule an apt? 

    Best,
    Dana
\end{quote}

Here we noticed a few differences to our Medical Q\&A example. 1) There is a sense that the doctor and patient know each other. 2) There is significant missing context from the patient --- i.e. How old are they? What is their sex? What pain are they referring to? What exact medication were they prescribed? What was the dosage? 3) The message is structured much differently. All of these stylistic qualities should be reflected in synthetic data used to train patient portal messaging systems. 

\subsection{Example Model Outputs}\label{apn:A_2}
In this section, we show the difference in generation performance between PortalGen and comparable LLM-prompting approaches. 

\paragraph{Prompt: } Patient is experiencing a diabetic coma and is unsure of what steps to take to regain control. 

\paragraph{GPT-3.5 Generation}
\begin{quote}
    Hey Doc, I'm in a bit of a tough spot right now. I'm experiencing a diabetic coma and I'm not sure how to regain control. I am dizzy, disoriented, and my blood sugar levels are all over. Can you give me some guidance on how to get back on track? 
\end{quote}

\paragraph{Mixtral 8x7b Generation}
\begin{quote}
    Hi Dr. I hope you are doing well, 

    I am very worried as I think I'm experiencing a diabetic coma. I have been feeling extremely dizzy and confused. I am also having difficulty staying conscious. 

    I'm not sure what to do and I am very scared. My blood sugar levels, which I have been monitoring, have been extremely high. My insulin does not seem to be helping. 

    What should I do? Should I go to the hospital or wait this out? I dont want to overreact, but I also don't know if I should ignore my symptoms. 

    Thank you
\end{quote}

\paragraph{PortalGen Generation}
\begin{quote}
    Hi Dr, 

    I am really scared. I just woke up, on the ground, and have no idea how I got there. I think I may have been in a diabetic coma. I am very nauseous and don't know what to do. I am worried I might slip into one again. Do I need to go to a hospital? My blood sugar has been very difficult to control. 

    Thanks,
\end{quote}

\paragraph{Qualitative Analysis: } We notice that in this example, the PortalGen output is much more natural and patient-like. Mixtral-8x7b's generation, conversely, is very formal and unrealistic for a patient experiencing this symptom. ChatGPT seems to attempt use of casual language, but also re-states the prompt in the message, producing an unnatural sounding output. 

One interesting limitation of Mixtral-8x7b is, across numerous prompts, Mixtral-8x7b relies on a fixed message format where there is a (i) message header (ii) description of symptom (iii) how they have been managing it (iv) request for help. While this can sometimes be appropriate, the models inability to generate messages of varying formats is not optimal.  

\section{Experimental Configurations}\label{apn:B}

\subsection{GPT-2 Fine-Tuned on Real Patient Data}
We use the GPT-2 fine-tuning script provided by \cite{dp-transformers} \footnote{\url{https://github.com/microsoft/dp-transformers/tree/main/examples/nlg-reddit/sample-level-dp}} to run our experiment. We fine-tune GPT-2 for 3 epochs on 1,000 real patient messages. We use the prefix `Patient Message: ' before all messages to provide a prompt the model can use for downstream data generation. We use the set of default hyperparameters for standard GPT-2 training from the cited repository. 

We then generate 1000 random patient messages using this prefix. Each synthetic message is a maximum of 256 tokens long. We use a temperature of 0.75 during generation. 

\subsection{GPT-2 Fine-Tuned with Differential Privacy (DP)}
We perform the same exact steps as for GPT-2 Fine-Tuned on Real Patient Data, but using the differential privacy training algorithm implementation from \cite{dp-transformers}. We use the set of default hyperparameters for DP-based GPT-2 training from the cited repository. 

\subsection{Zero-Shot LLM Prompting} 

We load each LLM in 4-bit quantized mode using BitsAndBytes \footnote{https://github.com/bitsandbytes-foundation/bitsandbytes}. We use Huggingface's \footnote{https://github.com/huggingface} text generation pipeline with temperature = 0.75 and max\_new\_tokens set to 256 for each generation. The prompt used is as follows: 

\begin{quote}
    Pretend you are a medical patient. Write a message to your doctor using the prompt: 

    \#\#\# Rules \#\#\# 
    
    - Assume the doctor you are messaging has been your physician for years. It is permissible to speak informally when appropriate. 
    
    - Do not restate the prompt in the message. 
    
    - You may add additional health context (e.g. symptoms or medications) to the message as needed. 

    Prompt: [prompt]

    Patient Message: 
\end{quote}

\subsection{PortalGen}

PortalGen uses the same prompting strategy and hyperparameters, but with 10 real messages in-context. We additionally write prompts for each of the 10 messages that could have in-theory generated them to teach the model how to map from prompt to target message. To promote output diversity, we shuffle the 10 messages in the prompt before each generation.  We additionally include a special ``\#\#\# End Of Message \#\#\#' token at the end of each exemplar prompt to encourage the model to write a single message and not a sequence of messages. 

\subsection{Training GPT-2 From Scratch}
For our perplexity-based experiments, we train a GPT-2 model from scratch using the 1000 synthetic messages from a given generation method. We then compute the perplexity of this model in the presence of 5000 real patient messages. In this setting, we train for GPT-2 for 10 epochs using the same hyperparameters as our standard GPT-2 fine-tuning experiment detailed above. 

\subsection{Additional Evaluation Details}

We note that throughout all LLM-based experiments, each model uses the exact same set of prompts to ensure fairness. Additionally, all 5,000 test messages used in quantitative evaluation are ensured to have character lengths between 500-1500 characters to ensure the message has sufficient content. 

\section{Stage \#1 Prompt Generation}\label{apn:c}

We use few-shot prompting to convert ICD-9 code descriptions into portal message prompts. The prompt used for this task is shown below: 

\begin{quote}
    Given an ICD9 code for a given patient, write a short description of a message that a patient might send to their doctor which may or may not be related to the code. Here are examples.

Example Code: [code description]

Example Message Description: 
\end{quote}

To obtain ICD-9 codes for generating patient message descriptions using this prompt, we randomly sample from a binned distribution of ICD-9 codes presented by patients in our dataset described in Section \ref{sec:data}. We bin the distribution of real ICD-9 codes into ICD-9 chapters (e.g. a single chapter would be ICD-9 codes 460-519: diseases of the respiratory system) to generally match a distribution of real patient health problems, then we sample individual codes (e.g. a single ICD-9 code would be 463: tonsillitis, acute) from each chapter uniformly at random. See the code supplement for the full prompt and sampling strategy used for this task.

\end{document}